%% file: main.tex
\documentclass[runningheads]{llncs}

\usepackage{eccv}

\usepackage{eccvabbrv}

\usepackage{graphicx}
\usepackage{booktabs}

\usepackage[accsupp]{axessibility}  

\usepackage{hyperref}

\usepackage{orcidlink}

\usepackage{multirow,multicol}
\usepackage{color}
\usepackage{soul}

\newcommand{\fk}[1]{\textcolor{black}{#1}}
\newcommand{\ffk}[1]{\textcolor{black}{#1}}
\newcommand{\yr}[1]{\textcolor{black}{#1}}

\newcommand{\tjs}[1]{\textcolor{black}{#1}}

\newcommand{\gjy}[1]{\textcolor{black}{#1}}

\begin{document}

\title{
FreeMotion: A Unified Framework for Number-free Text-to-Motion Synthesis} 

\titlerunning{Abbreviated paper title}


\author{Ke Fan\inst{1} \and
Junshu Tang\inst{1} \and
Weijian Cao\inst{2} \and
Ran Yi\inst{1} \and
Moran Li\inst{2} \and
Jingyu Gong\inst{1} \and
Jiangning Zhang\inst{2} \and
Yabiao Wang\inst{2} \and
Chengjie Wang\inst{1,2} \and
Lizhuang Ma\inst{1}
}

\authorrunning{F.~Author et al.}

\institute{Shanghai Jiao Tong University \and
Tencent Youtu Lab \\
\url{https://VankouF.github.io/FreeMotion}
}

\maketitle

\begin{abstract}
Text-to-motion synthesis is a crucial task in computer vision. Existing methods are limited in their universality, as they are tailored for single-person or two-person scenarios and can not 
be applied to generate motions for more individuals. To \yr{achieve} the number-free motion synthesis, this paper reconsiders motion generation and proposes to unify the single and multi-person motion by the \yr{conditional} motion distribution. Furthermore, a generation module and an \yr{interaction} module are designed for our \textit{FreeMotion} framework to decouple the process of conditional motion generation and finally support the number-free motion synthesis. Besides, based on our framework, the current single-person motion spatial control method could be seamlessly integrated, achieving precise control of multi-person motion. Extensive experiments demonstrate \yr{the superior performance of} our method \yr{and our capability to} infer single and multi-human motions simultaneously.
\end{abstract}

\input{sections/1_introduction}

\input{sections/2_related_work}

\input{sections/3_preliminary}

\input{sections/4_method}

\input{sections/5_experiment}
\input{sections/6_conclusion}


%
%
\bibliographystyle{splncs04}
\bibliography{main}
\end{document}

%% file: sections/1_introduction.tex
\section{Introduction}
\label{sec:intro}

\ffk{Human motion synthesis \yr{aims at} generating human motions driven by the input text, audio, scene, etc., and has extensive applications in areas like robotic control, movies, and animation production. \yr{Among these input signals,} text, {\it i.e.}, natural language, can provide rich semantic details and is the most user-friendly and convenient signal for motion synthesis. }

\ffk{Generating single-person motion, as a fundamental task in text-to-motion (T2M), has received significant attention in \yr{the} research \yr{community}. 
Some \yr{methods}~\cite{petrovich2022temos,guo2022generating} aligned the text and motion in the same space by utilizing the VAE structure, while others~\cite{zhang2023finemogen,zhang2023remodiffuse,xie2023towards} utilized the diffusion model conditioned on text to make the generated motion more diverse and realistic.}
\fk{Recently, some works have begun to explore \ffk{two-person} motion synthesis. ComMDM~\cite{shafir2023human} fine-tuned a pre-trained single-human motion generation model on a dataset with only 20 two-person text-annotated motions by few-shot learning. InterGen~\cite{liang2023intergen} further introduced a large annotated two-person motion dataset, InterHuman, and employed a shared-weight diffusion network and cross-attention to model the interaction of two-person motion.}

Although the aforementioned methods have demonstrated remarkable achievements in both single-person and \ffk{two-person} motion generation, their applicability remains limited in terms of universality. Specifically, these limitations include: (1) their models are tailored either for fitting the marginal motion distribution of single-person or the joint motion distribution of two-person scenarios and cannot support motion inference for both cases simultaneously. (2) Moreover, due to the \ffk{unavailability of text-annotated multi-person motion datasets}, 
\fk{as well as the \ffk{non-scalability of network design}}, none of these methods can be applied to generate motion for more than two individuals during the inference phase. (3) While some studies have accomplished spatial control for single-person motion generation, incorporating the spatial signal
into the current multi-person motion generation model remains a non-trivial challenge.

\ffk{To address these problems, we rethink the process of multi-person motion synthesis and propose FreeMotion, a novel unified framework} that enables motion generation for any number of individuals and achieves fine-grained spatial control of multiple human motions.

\begin{figure}[!t]
\centering
\includegraphics[width=0.9\linewidth]{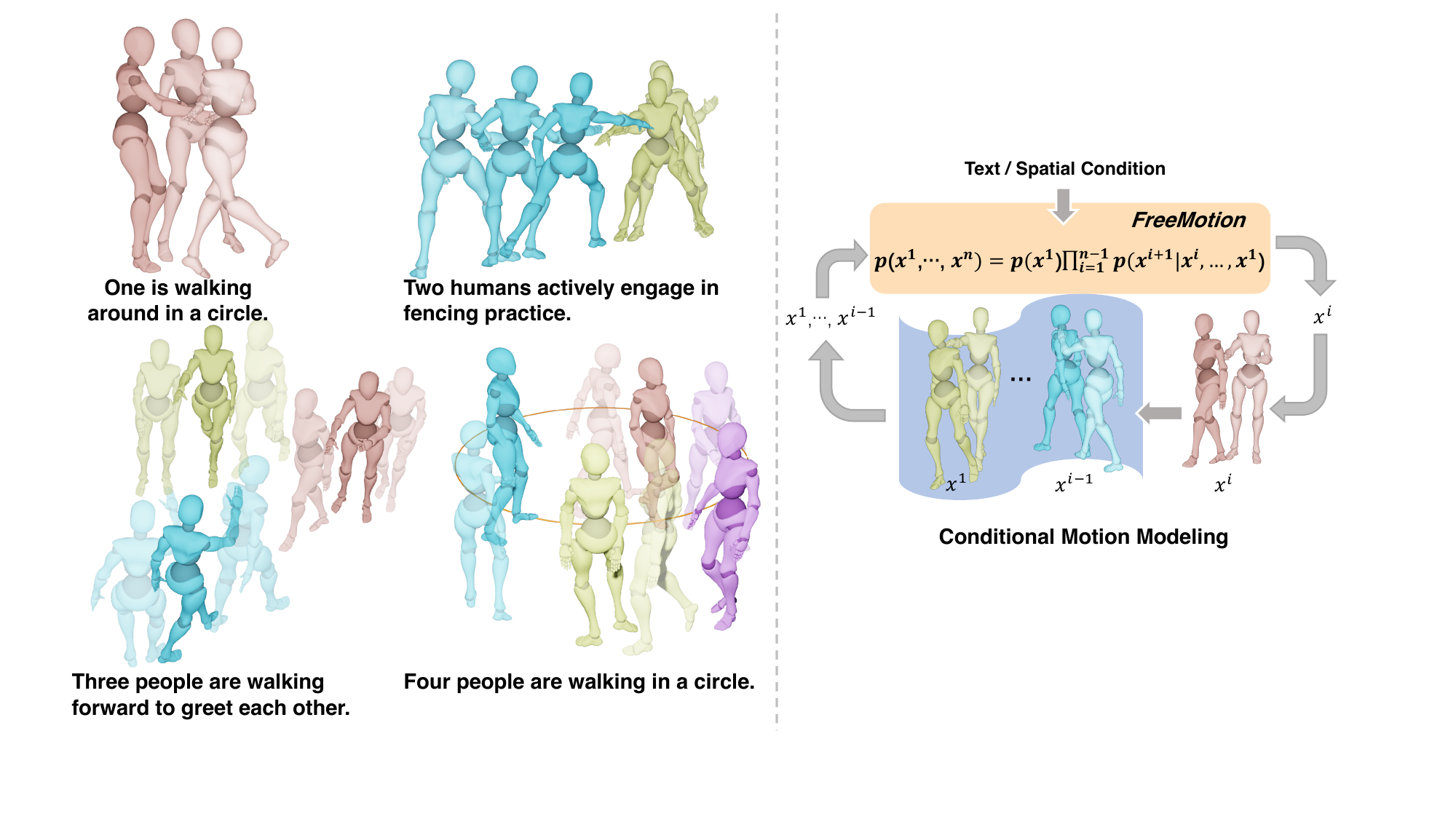}
\caption{The \textbf{left} shows our model can generate controllable motions for any number (1--4 from the figure) of individuals. Different colors represent the different person's motion. The \textbf{right} is an illustration of \yr{our} new paradigm of motion generation,
recursive generation, where every single motion is predicted under the condition of \yr{the} motion\yr{s} generated before. Best viewed in color.}
\label{motivation_fig}
\end{figure}

\ffk{Our key insight is to build a bridge between the marginal and joint distribution with the conditional distribution.} 
Without loss of generality, for the $n$-person motion generation task, 
we need to model the joint distribution probability $p(\mathbf{x}^1,...,\mathbf{x}^n)$. 
According to the formula of conditional probability, we can \yr{decompose} the joint distribution as $p(\mathbf{x}^1,...,\mathbf{x}^n) = p(\mathbf{x}^1)\prod_{i=1}^{n-1}p(\mathbf{x}^{i+1}|\mathbf{x}^{i},...,\mathbf{x}^1)$.
\yr{Therefore}, if we can model the conditional motion distribution, we can generate motions for $n$ individuals in a recursive process,
as shown in Fig.~\ref{motivation_fig}.
Through the conditional probability formula, we reduce the task of multi-person motion synthesis to the single-person motion synthesis under the guidance of other people's \yr{motions}, {\it i.e.}, conditional single-person motion synthesis.

To this end, \fk{we design a conditional motion diffusion network, which can generate the target motion conditioned on the motion\yr{s} of any number of other individuals.}
\tjs{We decouple the modeling of conditional motion distribution into single-person motion generation and multi-person motion interaction, which correspond to the two modules of our network, namely the generation module and the interaction module. 
The generation module aims to generate diverse and vivid single-person motion according to the text prompt.
\yr{While the interaction module aims to inject condition motions into the human motion generation by extracting the interactive information between condition motions and the current motion to be generated.}
To accommodate the variability in the number of motion conditions in the interaction module, we leverage the length-independent characteristics of self-attention and deliberately design the interactive block to be entirely based on global self-attention.}
\ffk{Besides, to describe the motion of each person,
we design prompts and employ a large language model~\cite{chatgpt} to transform multi-person motion descriptions into corresponding descriptions of each person.}

\tjs{
Furthermore, we enhance the fine-grained spatial control of multi-person motion generation. 
Previous work\yr{s}~\cite{wang2023intercontrol,xie2023omnicontrol} add accurate control on the joint of a single-person motion. Nevertheless, these control signals are not easy to be applied to multi-person motion generation. In our work, we incorporate flexible spatial signals into the interaction module \yr{to} control the global location of human motions while maintaining the realism of the motion result\yr{s}. 
As demonstrated \yr{in} Fig.~\ref{motivation_fig}, although our model is only trained on two-person motions, it
\yr{has} the ability to predict the movement\yr{s} of more than two individuals in the inference phase.}

In summary, our contributions are as follows: \ffk{(1) We rethink the process of motion synthesis and propose a new paradigm to unify the synthesis of motions for both single and multiple \yr{people}.}
(2) \yr{We} propose a decoupled generation and interaction module \yr{for conditional motion generation}, which \yr{is} the first attempt to achieve high-fidelity number-free motion generation under text conditions.
\ffk{(3)
We further achieve precise control of multi-person motion \yr{based on flexible spatial control signals utilizing explicit and implicit guidance}.} 
\yr{Extensive} experiments demonstrate that our method outperforms prior works, \yr{and achieves} vivid multiple motions result\yr{s}.

%% file: sections/2_related_work.tex
\section{Related Work}
\label{sec:rel_w}

\subsection{Single-Person Motion Synthesis} 
At present, the mainstream text2motion methods are mainly divided into two categories: align-based model and the condition-based model. 

The align-based model mainly aligns text and motion into a shared latent space. In the inference stage, features are extracted based on the given text, and the Features are regarded as corresponding motion features for action generation.
TEMOS ~\cite{petrovich2022temos} and Guo \textit{et al.} ~\cite{guo2022generating} leverage the VAE architecture to learn a joint latent space of motion and text constrained on a normal distribution.
However, natural language and human motions are quite different with misaligned structure and distribution, which makes the alignment process quite difficult.
 
Condition-based models are usually based on the diffusion model architecture. It uses pre-trained text encoders, such as CLIP, to extract text features, and inject text features as conditions into the diffusion reconstruction network to guide the network to generate corresponding motions.
MDM~\cite{tevet2022human} and Motion Diffuse~\cite{zhang2024motiondiffuse} are the first works to introduce the diffusion model into the motion generation field. 
MDM additionally introduces a geometric loss to improve the model performance.
MLD~\cite{chen2023executing} further leverages the latent diffusion model to significantly drop the training and inference cost.
However, Xie \textit{et al.}~\cite{xie2023towards} points out that the latent dimension affects the performance of the model. It cascades two diffusion models with different latent dimensions, promoting the details and the modal consistency.
ReMoDiffuse~\cite{zhang2023remodiffuse} proposes an enhancement mechanism based on data set retrieval to refine the denoising process of Diffusion. 
FineMoGen~\cite{zhang2023finemogen} and Motion-X~\cite{lin2023motion} further propose a new large-scale data set to introduce more detailed descriptions, such as hand joints, expression, etc. 
GMD~\cite{karunratanakul2023gmd} and OminiControl~\cite{xie2023omnicontrol} explore precise trajectory control by the impainting technique or both implicit and explicit spatial guidance.
However, these methods are all networks designed for single-person models, and it is difficult to achieve motion synthesis for two or even multiple individuals.

\subsection{Multi-Person Motion Synthesis} 
Multi-person motion synthesis is more difficult than single-human motion generation, which involves an interactive process between multiple individuals.
Early works tend to use motion graphs~\cite{shum2007simulating} and momentum-based inverse kinematics~\cite{komura2005animating}. 
Guo \textit{et al.}~\cite{guo2022multi} propose\yr{s} the Extreme Pose Interaction dataset as well as a two-stream network with cross-interaction attention for interaction modeling. 
InterFormer~\cite{chopin2023interaction} uses an attention-based interaction transformer to generate sparse-level reactive motions on the K3HI~\cite{hu2013efficient} and the DuetDance~\cite{kundu2020cross} datasets. 
SocialDiffusion~\cite{tanke2023social} proposes the first diffusion-based stochastic multiperson motion anticipation model. BiGraphDiff~\cite{chopin2024bipartite} further introduces bipartite graph diffusion for geometric constraints between skeleton nodes.
Tanaka \textit{et al.}~\cite{tanaka2023role}
introduces a PIT module the diffusion network, which enables the model to automatically distinguish actors and receivers, thereby better learning the interaction process. To broaden the applications, ReMos~\cite{ghosh2023remos} and Le \textit{et al.}~\cite{le2023music} further propose new datasets that contain hand movements or music.
However, these methods depend on either historical motion, action label, or music to give the motion prediction, and can not support the task of T2M.

ComMDM~\cite{shafir2023human} annotates 3DPW manually to obtain a small scale of samples. Furthermore, it fine-tunes a pre-trained single-person T2M model in a few-shot manner and attempted multi-person motion generation for the first time. However, since it only contains 27 two-person motion sequences, The model's ability to generate two-person interactions is greatly restricted. 
Therefore, InterGen~\cite{liang2023intergen} first contribute\yr{s} a large-scale text-annotated two-person motion dataset called InterHuman, and based on this, it proposed a diffusion model with shared weights and multiple regularization losses, outperforming the ComMDM.
InterControl~\cite{wang2023intercontrol} attempts to complete the interaction process by controlling the joints of two individuals to a certain position.
It uses LLM to generate planning for two individuals during their movements by designing prompts.
However, since it was not trained on two individuals, the model can not model the interaction process explicitly.
In short, these T2M models almost all directly predict the actions of a single person or a pair of individuals, and cannot support the inference of single and double persons at the same time, let alone support the inference of motion for multiple individuals at the same time.

\subsection{Diffusion Models}
Diffusion probabilistic models~\cite{ho2020denoising} are a class of neural generative models based on the stochastic diffusion process.
Compared to VAE-based pipelines, the most popular motion-generative models in previous works, diffusion models strengthen the generation capacity through a stochastic diffusion process, as evidenced by the diverse and high-fidelity generated results.
Some works have made great progress in accelerating the denoising process~\cite{song2020denoising} and improving the generation quality~\cite{nichol2021improved}.
Dhariwal \textit{et al.}~\cite{dhariwal2021diffusion}, introduce\yr{s} classifier-guided diffusion for a conditioned generation. The Classifier-Free Guidance approach~\cite{ho2022classifier} enables conditioning while trading-off fidelity and diversity, and achieves better results. In this paper, we implement our model condition\yr{ed} on the input text in a classifier-free manner.
ControlNet~\cite{zhang2023adding} copies the trainable parameters of a diffusion model to process the condition and freezes the original model to avoid degeneration of generation ability. Its generality in enabling various control signals has been largely proven. Inspired by this, we achieve conditional motion modeling by leveraging the ControlNet structure.

%% file: sections/3_preliminary.tex
\section{Preliminaries}

\subsection{Diffusion Model for Motion Synthesis}

The diffusion Model\cite{ho2020denoising} is a probabilistic model that gradually denoises a Gaussian noise to generate a target output. The key point is to generate a target output by gradually denoising Gaussian noise. It is formulated as a diffusion process and a reverse process and is utilized to approximate the posterior $q(\mathbf{x}_{1:T} \vert \mathbf{x}_0)$, where $T$ is the total time steps, and $\mathbf{x}_{1},...,\mathbf{x}_{T}$ are the real data $x_0$ with $t$ steps of noise added.
The diffusion process follows a Markov chain to gradually add Gaussian noise to the data until its distribution is close to the latent distribution $\mathcal{N}(\mathbf{0}, \mathbf{I})$, according to variance schedules given by $\beta_t$:
\begin{equation}
    \begin{aligned}
        &q(\mathbf{x}_{1:T} \vert \mathbf{x}_0) \,:=\, \prod_{t=1}^{T} q(\mathbf{x}_t \vert \mathbf{x}_{t-1}), \\
        &q(\mathbf{x}_t \vert \mathbf{x}_{t-1}) \,:=\, \mathcal{N}(\mathbf{x}_t; \sqrt{1-\beta_t}\mathbf{x}_{t-1}, \beta_t\mathbf{I}).
    \end{aligned}
\end{equation}

During training, the mean-squared error loss is used to optimize the parameters. 
During inference, we can generate the target output by sampling the initial noise $\epsilon$ and denoising it by predicting the added noise, mean, or $x_0$ recursively. To fit the arbitrary length of time, the transformer-based denoiser block is commonly preferred in motion generation.

\subsection{Motion Interaction Representation}
To maintain generality, we model multi-person motion generation as a motion sequence $\mathbf{X} = (\mathbf{x}^{p})$, $p \in \{1, 2, ..., N\}$, where $N$ represents the total number of individuals. 
For simplicity, we use $\mathbf{x}$ to represent the motion of a specific individual.

\ffk{In the context of generating motion for a single person, the canonical representation proposed by HumanML3D~\cite{guo2022generating} has proven to be well-suited for neural network processing. However, its normalization operation loses the relative position relationship in multi-person scenarios, \yr{making} it unsuitable for such use cases. To address this limitation, InterGen~\cite{liang2023intergen} introduced a non-canonical representation that maintains the relative positions between different individuals. The formulation of this representation is as follows:}
\begin{align}
x^p(i)=[\textbf{j}_{pg} , \textbf{j}_{gv} , \textbf{j}_r, \textbf{c}_f ],
\end{align}
where the $i$-th pose of $x^p$ is defined as a collection of global joint positions $\textbf{j}_{gp}\in \mathbb{R}^{3J}$, velocities $\textbf{j}_{gv}\in \mathbb{R}^{3J}$ in the world frame, local rotations $\textbf{j}_{r}\in \mathbb{R}^{6J}$ in the root frame, and binary foot-ground contact features 
$\textbf{c}_{f}\in \mathbb{R}^4$, where $J$ denotes the joint number.
Since this representation preserves the global coordinates of joints, it well represents the spatial relationship of interactions. In this paper, we follow the non-canonical representation proposed by InterGen.

%% file: sections/4_method.tex
\section{Method}

\subsection{Overview}
\tjs{In this paper, we propose a unified framework for number-free human motion generation. By reconsidering the generation process via decomposing the joint motion into a conditional distribution through the conditional probability formula, our framework can generate arbitrary number of motion\yr{s} in a recursive manner, as shown in Fig.~\ref{motivation_fig}.}
Specifically, we model the multi-person joint motion distribution as $p(\mathbf{x}^1,...,\mathbf{x}^n)\\=p(\mathbf{x}^1)\prod_{i=1}^{n-1}p(\mathbf{x}^{i+1}|\mathbf{x}^i,...,\mathbf{x}^1)$. 
Given a text $\mathbf{d}$, we \yr{first predict} the motion of the first individual. Subsequently, the motion of the second individual is inferred, condition\yr{ed} upon the motion of the first individual, and the process iterates until the motion of the \textbf{n}-th individual is generated.
Therefore, this method essentially transforms a multi-person motion generation process into \yr{a sequence of} conditional single-person motion generation. Based on this, we can further achieve multi-human spatial motion control by separately controlling the spatial motion of each individual.

\subsection{Number-free Motion Generation}

\begin{figure}[!t]
\centering
\includegraphics[width=\linewidth]{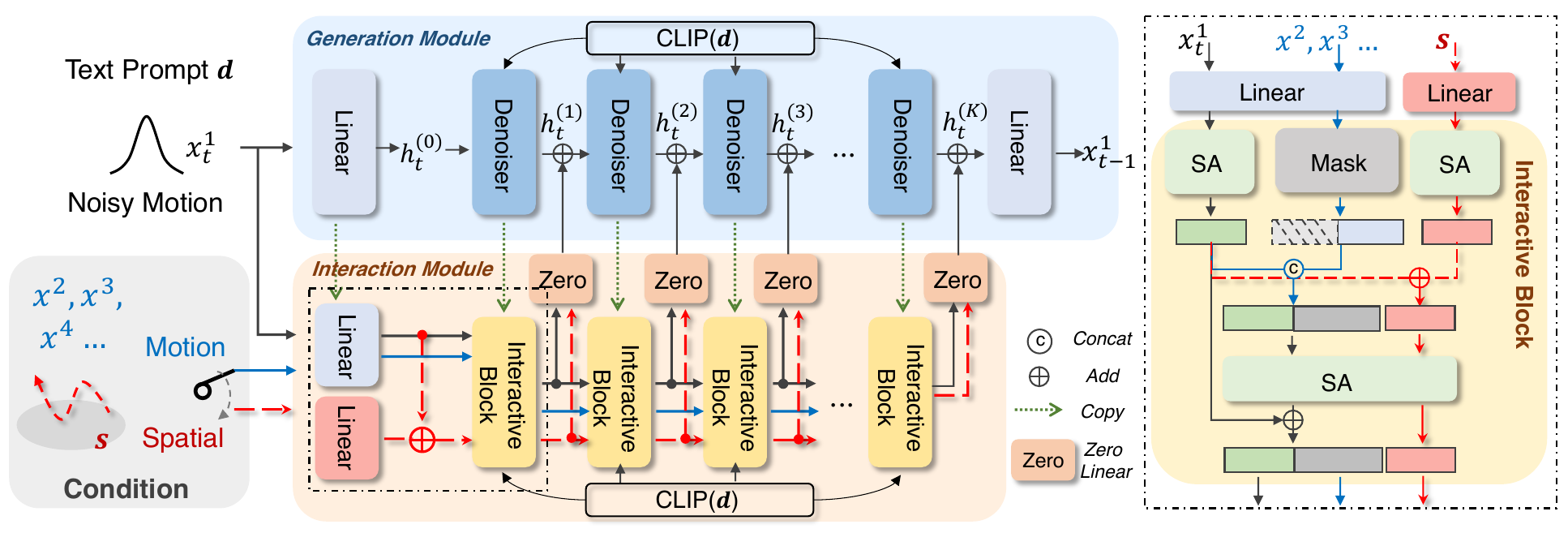}
\caption{
\tjs{Overall architecture of FreeMotion, which contains a generation module and an interaction module. Given a text $\mathbf{d}$, 
our framework can infer a motion $x^{1}$ by the generation module independently, or under the condition of multiple motions $x^{2}, x^{3}...$ or some spatial guidance $\mathbf{s}$. Red line represents the implicit guidance of the spatial control signal.}
}
\label{model}
\vspace{-20pt}
\end{figure}

\tjs{
As introduced before, we consider the multiple motions generation as a conditional generation process. \ffk{From the perspective of conditional motion distribution, we decouple the modeling process into a motion generation process and a motion interaction process.} To start with, we first design a generation module which has the capability of single motion generation. Then we design an interaction module
\yr{to inject condition signals (motions of $N-1$ individuals) into the human motion generation process by modeling the interaction between condition signals and the current motion to be generated.}
The detailed architecture of 
the generation and the interaction module are shown in Fig.~\ref{model}.
}

\textbf{Generation Module.} The generation module is designed to \yr{synthesize} diverse single motion according to the text prompt. Similar \yr{to} previous text-to-motion methods~\cite{zhang2024motiondiffuse,tevet2022human}, the module is a Transformer-based diffusion network, containing several denoiser blocks. During training, at timestep $t$, we add noise to \yr{a} motion $\mathbf{x}$ and get the \yr{noised motion} $\mathbf{x}_{t}$, \yr{and} then use the generation module to denoise $\mathbf{x}_t$ to $\mathbf{x}_{t-1}$. During the inference, the generation module is able to generate a clean motion $x_{0}$ starting from the pure Gaussian noise $\mathbf{x}_{T}$. 

\textbf{Interaction Module.} 
Inspired by ControlNet~\cite{zhang2023adding}, we \yr{design} a neural network named Interaction Module to inject the condition signals into the human motion generation. 
\tjs{
Specifically, \yr{we denote the} $N-1$ motion\yr{s in the condition signal as} $\mathbf{x}^2, \mathbf{x}^3, .., \mathbf{x}^N$, 
\yr{and} the noised motion \yr{obtained by adding noise to $\mathbf{x}^1$ as} $\mathbf{x_t}^1$. 
\yr{Firstly, we feed $\mathbf{x}_t^1$} and \yr{$N-1$} motion \yr{conditions} $\mathbf{x}^i, i\in\{2,...,N\}$ \yr{into}
a shared linear layer \yr{to} encode \yr{each motion in}to a hidden state, \ffk{denoted as $\mathbf{h}_t^{1,0}$ and $\mathbf{h}^{i,0}, i \in \{2,...N\}$.}}
Secondly, to model the interaction information between variable number of motion conditions 
\yr{and the motion to be generated,}
we design a novel \textbf{Interactive Block}, which is sequentially stacked for $K$ times. 
\ffk{The interactive block at the $k$-th stage ($k \in \{1,...,K\}$) takes the $\mathbf{h}_t^{1,{k-1}}$ and $\mathbf{h}^{i,{k-1}}$ \yr{as inputs} and outputs the $\mathbf{h}_t^{1,{k}}$ and $\mathbf{h}^{i,{k}}$}.
Each \yr{interactive} block contains two sequential self-attention modules and a mask module, \ffk{and the whole \yr{calculation} process in the \yr{$k$-th interactive} block is formulated as: }
\begin{align}
&\mathbf{h}^{1,{k}}_t, \mathbf{h}^{2,{k}},...,\mathbf{h}^{N,{k}} = SA( SA(\mathbf{h}^{1,{k-1}}_t), Mask(\mathbf{h}^{2,{k-1}},...,\mathbf{h}^{N,{k-1}})),
\end{align}
\ffk{where $k\in \{1,..., K\}$ and $SA$ represents the self-attention operation.}

\ffk{Specifically, in the $k$-th interactive block, we firstly randomly mask some hidden states $\mathbf{h}^{i,k-1}$, to \yr{enable} our model \yr{to} infer the motion under any number of motion conditions. 
\yr{Then, t}he noised \yr{hidden state} $\mathbf{h}^{1,k-1}$ \yr{is} input into the first $SA$ module.}
\tjs{
We concatenate the masked $N-1$ \ffk{hidden states}, and the output of the first $SA$ module along the channel dimension, and feed the concatenated result to the \ffk{second $SA$ module, so that the model learns the interactive information between the noised motion and motion conditions. And the outputs \yr{of the second $SA$ module, {\it i.e.,}} $\mathbf{h}_t^{1,k}$ and $\mathbf{h}^{i,k}$s, will be further input to the next interactive block. By leveraging the global self-attention, $\mathbf{h}_t^{1,k}$ successfully fuses the \yr{information from the motion conditions}.}} 

\yr{The output from the interactive block} will \yr{then} be added to the output of each denoiser block in the generation module after going through a linear layer. The linear layers are initialized with zero, thereby ensuring that the interactive module does not impact the generation module at the onset of training. Compare\yr{d} with cross-attention, self-attention is more flexible for modeling the interaction between motions regardless of the number of motion\yr{s in the} condition.

\subsection{Training Process}

\ffk{We \yr{decompose} our training process into two stages.}
\yr{In the first stage, w}e train the generation module for single-person motion generation.
To enable the generation module to synthesize the single-person motion, it is necessary to \yr{provide} a single-person motion description \yr{as its input} during training. 
Therefore, for a given interactive description of multi-person motion, we utilize the ability of the Large Language Model (LLM)~\cite{chatgpt} to generate $N$ motion description\yr{s, one} for each individual.
We use the separated single-person description for the training of the generation module. The text feature is extracted by a pre-trained CLIP~\cite{radford2021learning} model and is injected \yr{in}to the adaptive layer normalization in all attention layers.

\ffk{In the second stage, we train the interaction module for conditional motion modeling and multi-human motion generation.
} 
\ffk{Inspired by the success of ControlNet, we first freeze the parameters of the generation module and utilize the parameters to initialize the interaction module.} 
\tjs{Then we use multi-human motions \yr{data to} train the interaction module.}
\yr{Denote} the motion\yr{s} of $N$ individuals as
$[\mathbf{x}^1,\mathbf{x}^2,...,\mathbf{x}^N]$.
We initially randomize the order of these $N$ motions and \yr{the} corresponding descriptions, \yr{and} then select the first one as the motion to be noised and reconstructed, leaving the other \yr{$N-1$ motions} as the clean motion condition. 
\yr{1)} We add $t$ steps of noise to $\mathbf{x}^1$ to obtain $\mathbf{x}^1_t$ and input it to the generation module. 
\yr{2)} For the interaction module, $\mathbf{x}^1_t$ and the remaining $N-1$ motion conditions $\mathbf{x}^p$, $p\in \{2,..., N\}$ are \ffk{input into a shared linear layer and $K$ interactive blocks to extract the interactive information. 
\yr{3)} The output of each interactive block is added to the output from the corresponding denoiser block \yr{in the generation module} after going through a linear layer.
\yr{4)} Finally, the denoised motion $x^1_{t-1}$ can be obtained from the generation module. }

\subsection{Spatial Control}
Our proposed framework \yr{based on conditional motion generation} enables effortless spatial control over multi-human motions. It allows for the integration of the current single-person motion control method without the need for carefully designing the spatial representation of multiple individuals. Inspired by OmniControl~\cite{xie2023omnicontrol}, we concurrently utilize explicit and implicit guidance to realize our spatial control.

\textbf{Explicit guidance.} 
Given a desired spatial location $\mathbf{s} \in R^{F\times 3J}$, where $F$ represents the valid motion length and $J$ denotes the joint number. We \yr{utilize} the $L_2$ distance $\mathbf{d}=\left \| \mathbf{s}_{nj} - \mathbf{x}_{nj}  \right \|_2$ to measure the bias between target position $s$ and the predicted motion $\mathbf{x}$ for joint $j$ at frame $n$. 
Then we leverage the mechanism of classifier guidance
to perturb the predicted motion at each denoising step $t$ to \yr{correct} the bias, which is formulated as $\mathbf{x}_t = \mathbf{x}_t - \eta \nabla_{\mathbf{x}_t} \mathbf{d}\label{eq:x_t_perturb}$, where $\eta$ controls the step of the guidance.

\textbf{Implicit guidance.} 
In addition to explicit guidance, as the red lines shown in Fig.~\ref{model}, we further employ implicit spatial guidance. The whole training process is almost the same as the number-free motion generation. A minor difference is that we first input the spatial signal $\textbf{s}$ into an independent linear layer, and add the output together with the noised motion \yr{hidden state}. Some frames and joints are randomly selected and the others are subsequently masked out for the whole training process.

\tjs{After injecting the spatial control signal under explicit and implicit guidance, our proposed unified framework has the capability to control the spatial location\yr{s} of multi-human motions independently, which further enhances the controllability of our method.}

\subsection{Loss Function} 

In addition to reconstruction loss $\mathcal L_{rec}$, we also incorporate some regularization losses, which
include the contact loss $\mathcal L_{foot}$ and joint velocity loss $\mathcal L_{vel}$ mentioned in MDM~\cite{tevet2022human} and the bone length loss $\mathcal L_{bl}$
\ffk{ as well as the masked joint distance map (DM) loss $\mathcal L_{dm}$}
mentioned in InterGen~\cite{liang2023intergen}.  
For the first stage of single-motion generation, the total loss is \yr{formulated} as: 
\begin{align}
\mathcal{L}_1 =\mathcal{L}_{rec}+\lambda_1 \mathcal{L}_{foot}+\lambda_2 \mathcal{L}_{vel}+\lambda_3 \mathcal{L}_{bl},
\end{align}
where \yr{the} DM loss is not included. For the second stage of conditional single-motion generation, we further introduce DM loss to model the interactions, \yr{where the total loss} is formulated as: 
\begin{align}
\mathcal{L}_2 =\mathcal{L}_{rec}+\lambda_1 \mathcal{L}_{foot}+\lambda_2 \mathcal{L}_{vel}+\lambda_3 \mathcal{L}_{bl}+\lambda_4 \mathcal{L}_{dm},
\end{align}
\yr{where} $\lambda_1,\lambda_2,\lambda_3$, and $\lambda_4$ are used to balance the effect of corresponding loss.

%% file: sections/5_experiment.tex
\section{Experiments}

\subsection{Datasets and Metrics}
\textbf{Datasets.} We evaluate our proposed framework \gjy{on} the \textbf{InterHuman} dataset, \yr{which} is the first text-annotated two-person motion dataset, consisting of 6,022 motions derived from various categories of human actions, labeled with 16,756 unique descriptions composed of 5,656 distinct words.

\noindent \textbf{Metrics.} We \yr{utilize} the same evaluation metrics as InterGen. (1) FID: measuring the latent \gjy{distribution} distance between the generated dataset and the real dataset. (2) R Precision: measuring the text motion matching, indicates the probability that the real text appears in the Top-k (1, 2, and 3) after sorting. (3) Diversity: measuring latent variance. (4) Multimodality (MModality): measuring diversity within the same text. (5) Multi-modal distance (MM Dist): measuring the distance between motions and text features.

\subsection{Implementation Details}
We employ a frozen \textit{CLIP-ViT-L-14} model as the text encoder. The number of diffusion timesteps is set to 1,000 during training and the DDIM~\cite{song2020denoising}  sampling strategy with 50 timesteps is applied \gjy{in the inference stage}. For \textbf{number-free motion generation}, the batch size is set \yr{to} 80 and 30 for the first and the second training stage on each GPU. The epoch \yr{number}s are 2,500 and 1,000 separately for the two stages. Both training stages are trained with 1e-4 learning rate and 2e-5 weight decay. For \textbf{spatial control}, the epoch \yr{number} and learning rate are set \gjy{to} 1,000 and 1e-5. Other hyper-parameters \gjy{remain} the same as aforementioned. All experiments are trained on eight Tesla V100 GPUs.

\subsection{Quantitative Results}
\textbf{Baseline.} \fk{We mainly compare our method against the current state-of-the-art approach \textit{InterGen} that models the interaction relationship of two \yr{people} by a cross-attention mechanism. However, it \yr{by default} can only support the two-person motion inference \yr{and cannot generate single-person motion}.}
\gjy{Thus, we make minor modifications to its model structure and retrain it to make it adapt to the inference of both single and double human motion, ensuring a fair comparison.} 
The corresponding output of cross-attention is set to zero with a 10\% probability, thereby gaining some capacity for single-person generation. Our re-annotated single-person texts are utilized during the training phase. \ffk{Since the performance of other methods is inferior to the InterGen on two-person motion synthesis, we directly take the results reported by InterGen for the experiments of two-person motion synthesis.}

\input{tabs/two_human_results}

\noindent \textbf{Comparison Results.} \gjy{For evaluation on two-person motion synthesis, we follow the settings given by InterGen. As for single-person motion synthesis, we take the corresponding re-annotated two single-person descriptions given by LLM to synthesize two separate single-person motions and concatenate them together for performance evaluation.}  
It can be seen from Tab.~\ref{two_human_comparison} and Tab.~\ref{single_human_comparison}  that although the InterGen* can \yr{generate} both single and two-person motions, the performance is \gjy{inferior to ours in almost all terms of metrics.}
Besides, the remaining models in Tab.~\ref{two_human_comparison} are only trained on the interactive descriptions and have no ability for single-motion inference, their performance on two-person motion synthesis is still inferior to our method. This fully demonstrates the powerful generality of our proposed framework.

\input{tabs/single_human_results}

\subsection{Ablation Studies}
We investigate the influence of several designs on two-person motion generation. Initially, we remove the interaction module, solely employing the generation module (GM). During the second training stage, we finetune the parameters of GM. To endow the network with both single-person and two-person motion capabilities, we drop the motion condition with a 10\% probability in the second SA module within each denoiser block. In addition, we exclusively use our re-annotated single-person motion texts, without the interactive description (InterDes) provided in the InterHuman dataset. Subsequently, we incorporate high-quality two-person interactive motion descriptions (2nd row) during the second training stage. 
Furthermore, we introduce the interaction module as well as InterDes to get our FreeMotion* (3rd \yr{row}) and FreeMotion (4th row).

Compar\yr{ing} FreeMotion* with GM*, we found it almost gets the same results in all metrics. However, after introducing the high-quality interactive description (InterDes), our FineMotion achieved a significant improvement while the performance of GM dropped a great deal.
We analyzed this \yr{and found that it is} because \yr{when} using only GM, the GM network parameters need to be updated in the second training stage. When using only single-person text, the text used in the second stage of training was consistent with the first stage and did not produce a large gap, so that the update process was able to ensure that GM* learned as well as FineMotion*. However, when InterDes was introduced, the differences between single and double text prevented the GM from effectively adapting its parameters after the first stage of training. However, as can be seen from the FineMotion results, the introduction of high-quality two-person texts is of great significance in improving the model performance. Overall, the decoupled generation module and interaction module we designed can utilize all the information for training more efficiently.

\input{tabs/ablation1}

\subsection{Qualitative Results}

\begin{figure}[!t]
\centering
\includegraphics[width=0.9\linewidth]{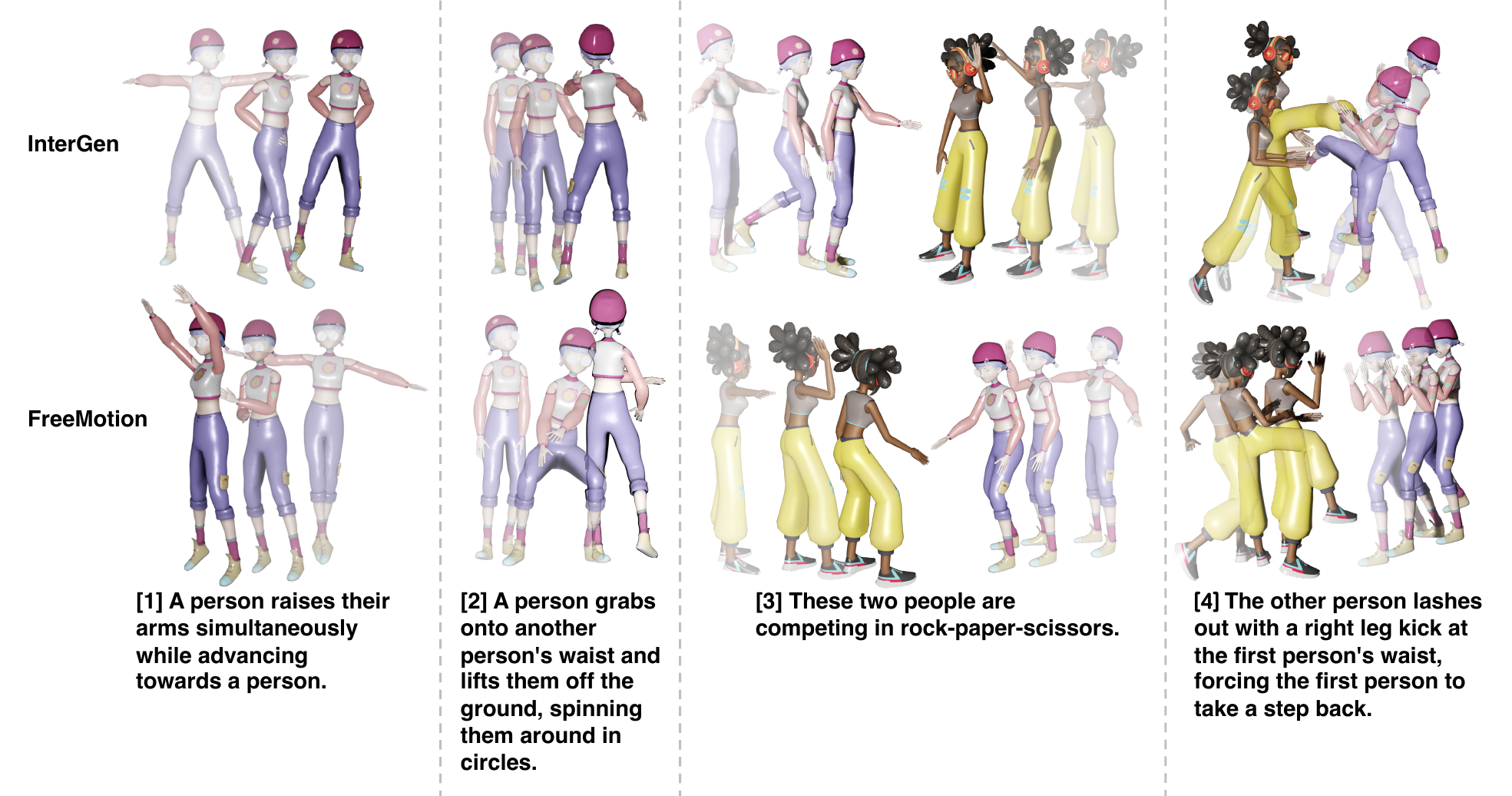}
\caption{Comparison with Intergen* on single and two-person motion generation. For single-person motion, we generate it with our re-annotated single description. For two-person motion, we further leverage the original interactive descriptions. For better visualization, some pose frames are shifted to prevent complete overlap.}
\label{vis_onetwo}
\vspace{-10pt}
\end{figure}

\textbf{Single Person Motion Synthesis.}
To illustrate the effectiveness, we provide a qualitative comparison between the \gjy{InterGen$^*$}
and our FreeMotion on single-motion synthesis. As shown in Fig.~\ref{vis_onetwo}, the \gjy{synthesized single-person motion given by the proposed method are more consistent with the description.} 
For example, in the second column, our model can produce an obvious squatting pose, which is more consistent with the textual command of grabbing others' waists. This fully demonstrates that the decomposition of the generation and interaction process we proposed can more adequately fit the mapping from text to single motion than randomly mask the interaction process in the training phase. 

\noindent \textbf{Two Person Motion Synthesis.} We further exhibit the comparison of two\gjy{-person} motion synthesis in the 3rd and 4th columns in Fig.~\ref{vis_onetwo}. 
\begin{figure}[!tbh]
\centering
\includegraphics[width=0.8\linewidth]{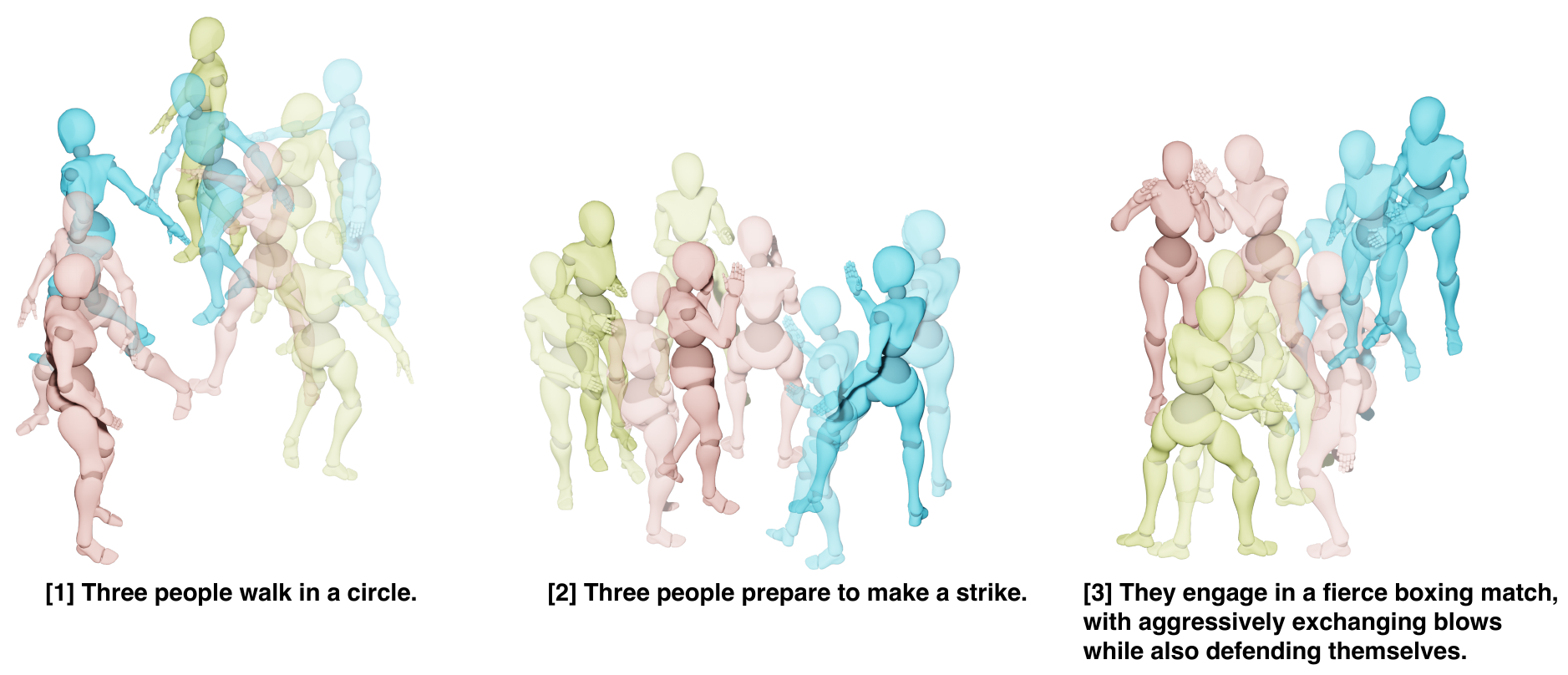}
\caption{Qualitative results for generating three-person motions. We manually design some text prompts and feed them to our network for motion generation. For better visualization, some pose frames are slightly shifted to prevent completed overlap.}
\label{vis_three}
\vspace{-20pt}
\end{figure}
Our approach performs better in two-person interaction coordination and comprehensibility in complex text. It can be seen that \gjy{InterGen$^*$} does not achieve simultaneity 
when generating the rock-paper-scissors motion at the third column, and even switches the mainly used hands. In the fourth column, \gjy{InterGen$^*$} only focuses on the kicking motion, which results in both people making kicks. In contrast, our method can perform better. We believe that this is also due to the decoupling of the generation and interaction processes. The generation module makes the comprehension of complex text better, and the interaction module enables better two-person \gjy{synchronization.}

\noindent \textbf{Three Person Motion Synthesis.}  
We manually designed the text with three different levels of interaction (low, middle, and high), corresponding to the results from left to right in Fig.~\ref{vis_three}.
Our interactive block is designed with global self-attention, which allows our network to have the ability to support more people's motions as conditions, thus generating three people's motions. More importantly, although our approach is trained using only two people's motions due to data scarcity, the way we decouple the generation and interactive processes enables our network to have the ability to directly reason about three people's motions. The generation module enables our network to generate semantically consistent one-person motions under weak interaction text. The interaction module enables the extraction of motion interaction features under strong interaction text. The above results further illustrate that the interactive block we designed can extract interaction information effectively.

\begin{figure}[!t]
\centering
\includegraphics[width=0.9\linewidth]{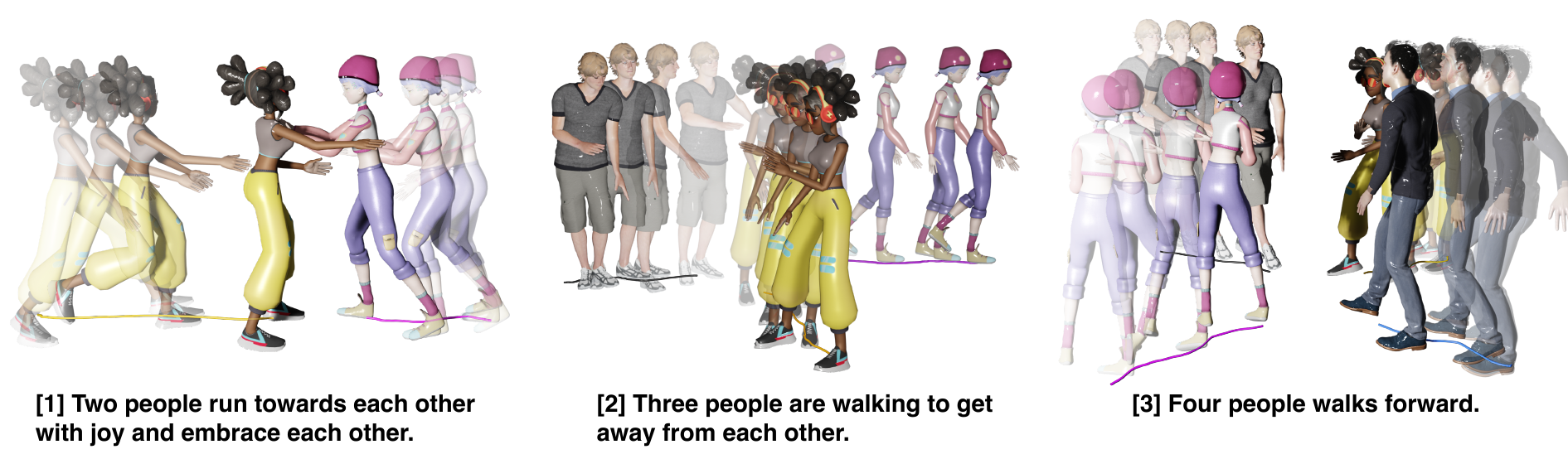}
\caption{Results of multi-person spatial control. We manually design some text prompts as well as the trajectories and leverage the integrated spatial control module to generate the results.}
\label{vis_trajectory}
\vspace{-20pt}
\end{figure}

\noindent \textbf{Controllable Motion Synthesis.}  Thanks to our proposed paradigm of conditional motion modeling, we can seamlessly integrate existing single motion control models \gjy{into} spatial control of multi-person motions. We use both implicit and explicit spatial guidance to control our generation module (i.e., controlling the single-person motion generation). As shown in Fig~\ref{vis_trajectory}, even though the spatial control module is trained for single motions, mounting the interaction module when generating multi-person motions has no obvious damage to the spatial control and performs well from two to four-person.

%% file: tabs/two_human_results.tex
\begin{table}[tbp]
	\begin{center}
		\centering
    \caption{Quantitative comparisons on the InterHuman test set.  We run all the evaluations 20 times except MModality runs 5 times. $\pm$ indicates the 95\% confidence interval. Bold indicates the best result.}
        \label{two_human_comparison}
		
 		\resizebox{1\textwidth}{!}{
        		\begin{tabular}{lccccccc}
        		    \toprule
        		    \multirow{2}{*}{Methods}  & \multicolumn{3}{c}{R Precision$\uparrow$} & \multirow{2}{*}{FID $\downarrow$} & \multirow{2}{*}{MM Dist$\downarrow$}  & \multirow{2}{*}{Diversity$\rightarrow $} & \multirow{2}{*}{MModality $\uparrow$}\\
        		    \cmidrule(lr){2-4}
        			& Top 1 & Top 2  & Top 3 \\
        			\midrule
                        Real    &  $0.452^{\pm .008}$ &  $0.610^{\pm .009}$  & $0.701^{\pm .008}$   &  $0.273^{\pm .007}$ &  $3.755^{\pm .008}$  & $7.748^{\pm .064}$   & -  \\
                        \midrule
        			TEMOS {\cite{petrovich2022temos}}  &  $0.224^{\pm .010}$ &  $0.316^{\pm .013}$  & $0.450^{\pm .018}$   &  $17.375^{\pm .043}$ &  $5.342^{\pm .015}$  & $6.939^{\pm .071}$   & $0.535^{\pm .014}$  \\
        			T2M {\cite{guo2022generating}}  &  $0.238^{\pm .012}$ &  $0.325^{\pm .010}$  & $0.464^{\pm .014}$   &  $13.769^{\pm .072}$ &  $4.731^{\pm .013}$  & $7.046^{\pm .022}$   & $1.387^{\pm .076}$ \\
        			MDM {\cite{tevet2022human}}   &  $0.153^{\pm .012}$ &  $0.260^{\pm .009}$  & $0.339^{\pm .012}$   &  $9.167^{\pm .056}$ &  $6.125^{\pm .018}$  & $7.602^{\pm .045}$   & $\textbf{2.355}^{\pm .080}$ \\
                    ComMDM$^*$ {\cite{shafir2023human}}  &  $0.067^{\pm .013}$ &  $0.125^{\pm .018}$  & $0.184^{\pm .015}$   &  $38.643^{\pm .098}$ &  $13.211^{\pm .013}$  & $3.520^{\pm .058}$   & $0.217^{\pm .018}$  \\
                    ComMDM {\cite{shafir2023human}}  &  $0.223^{\pm .009}$ &  $0.334^{\pm .008}$  & $0.466^{\pm .010}$   &  $7.069^{\pm .054}$ &  $5.212^{\pm .021}$  & $7.244^{\pm .038}$   &  $1.822^{\pm .052}$ \\
        			
                    InterGen$^*$ {\cite{liang2023intergen}}   &  $0.264^{\pm .006}$ &  $0.392^{\pm .005}$  & $0.472^{\pm .005}$   &  $13.404^{\pm .200}$ &  $3.882^{\pm .001}$  & $7.77^{\pm .030}$ & $1.451^{\pm .034}$ \\
                    \midrule
                    FreeMotion  &  $\textbf{0.326}^{\pm .003}$ &  $\textbf{0.462}^{\pm .006}$  & $\textbf{0.544}^{\pm .006}$   & $\textbf{6.740}^{\pm .130}$ &  $\textbf{3.848}^{\pm .002}$  & $\textbf{7.828}^{\pm .130}$ & $1.226^{\pm .046}$ \\
                    \bottomrule
        		\end{tabular}
 		}
	\end{center}

\end{table}

%% file: tabs/single_human_results.tex
\begin{table}[tbp]
	\begin{center}
		\centering
		\caption{Quantitative comparisons with InterGen* on our re-annotated text for single motion generation. The manner of evaluation is the same as Tab.~\ref{two_human_comparison}. Bold indicates the best result.}
        \label{single_human_comparison}
		
 		\resizebox{1\textwidth}{!}{
        		\begin{tabular}{lccccccc}
        		    \toprule
        		    \multirow{2}{*}{Methods}  & \multicolumn{3}{c}{R Precision$\uparrow$} & \multirow{2}{*}{FID $\downarrow$} & \multirow{2}{*}{MM Dist$\downarrow$}  & \multirow{2}{*}{Diversity$\rightarrow $} & \multirow{2}{*}{MModality $\uparrow$}\\
        		    \cmidrule(lr){2-4}
        			& Top 1 & Top 2  & Top 3 \\
        			\midrule
                        Real    &  $0.452^{\pm .008}$ &  $0.610^{\pm .009}$  & $0.701^{\pm .008}$   &  $0.273^{\pm .007}$ &  $3.755^{\pm .008}$  & $7.748^{\pm .064}$   & - \\
                        \midrule
        			InterGen*{\cite{liang2023intergen}}  &  $0.206^{\pm .004}$ &  $0.313^{\pm .004}$  & $0.389^{\pm .005}$   &  $23.415^{\pm .222}$ &  $3.925^{\pm .001}$  & $7.514^{\pm .029}$   & $\textbf{1.526}^{\pm .026}$  \\
                    \midrule
                    FreeMotion    &  $\textbf{0.264}^{\pm .005}$ &  $\textbf{0.394}^{\pm .006}$  & $\textbf{0.473}^{\pm .006}$   &  $\textbf{12.975}^{\pm .171}$ &  $\textbf{3.885}^{\pm .035}$  & $\textbf{7.702}^{\pm .027}$ & $1.300^{\pm .063}$ \\
                    \bottomrule
        		\end{tabular}
 		}
		\vspace{-20pt}
	\end{center}

\end{table}

%% file: tabs/ablation1.tex
\begin{table}[tbp]
	\begin{center}
		\centering
		\caption{Ablation study of our proposed framework. All results are reported on the InterHuman test set under the setting of two-person motion generation. \textbf{InterDes} and the \textbf{GM} represent the interactive description and generation module. We use \* to separate whether to leverage InterDes or not.} 
        \label{ablation1}
		
 		\resizebox{1\textwidth}{!}{
        		\begin{tabular}{lccccccc}
        		    \toprule
        		    {Methods}  & InterDes & {R-Precision1$\uparrow$} & {FID $\downarrow$} & {MM Dist$\downarrow$}  & {Diversity$\rightarrow $} \\
                        \midrule
        			{GM*}  &  & $0.300^{\pm .005}$ &  $8.842^{\pm .130}$ &  $3.863^{\pm .001}$  & $7.761^{\pm .036}$    \\
        			{GM}   & \checkmark &
                    $0.259^{\pm .005}$ &  $10.749^{\pm .145}$ &  $3.883^{\pm .001}$  & $7.645^{\pm .031}$ \\
        			FreeMotion{*}   & &  $0.300^{\pm .004}$ &  $8.792^{\pm .135}$ &  $3.865^{\pm .001}$  & $7.750^{\pm .022}$  \\
                    
                    \midrule
                    FreeMotion    & \checkmark &  $\textbf{0.326}^{\pm .003}$ & $\textbf{6.740}^{\pm .130}$ &  $\textbf{3.848}^{\pm .002}$  & $\textbf{7.828}^{\pm .130}$  \\
                    \bottomrule
        		\end{tabular}
 		}
		\vspace{-20pt}
	\end{center}

\end{table}

%% file: sections/6_conclusion.tex
\section{Conclusion and Limitations}
\vspace{-5pt}
In this paper, we present FreeMotion, a novel motion generation framework for number-free motion synthesis. We rethink the way of multi-person motion generations and propose to recursively generate multi-person motions based on conditional motion modeling. We further propose the decoupled generation module and interaction module, and conduct a large number of quantitative as well as qualitative experiments to prove that our framework can support motion synthesis for any number of motions.

\noindent{\textbf{Limitations.}} First, we directly utilize the capability of the large language model to separate single-motion text from the interactive description. Inevitably, there might be some instances where text and movements do not match well, which limits our ability to generate single-person motion.
Secondly, although our model can generate multi-person motions, due to the limited understanding of interactions from training only on two-person motions, 
there might be some interpenetration among different individuals when the number gets large or the text prompt is too complex.